\documentclass[11pt, a4paper, logo, copyright]{googlecloud}

\usepackage[authoryear, sort&compress, round]{natbib}

\hypersetup{
  colorlinks,
  citecolor=blue,
  linkcolor=red,
  urlcolor=blue}
\usepackage{nicefrac}
\usepackage{wrapfig}
\usepackage{soul}
\usepackage{adjustbox}
\usepackage{arydshln}
\usepackage[capitalize,noabbrev]{cleveref}

\usepackage{inconsolata}
\usepackage[skins,breakable,most]{tcolorbox}

\usepackage{subcaption}
\usepackage{algorithm}
\usepackage{algpseudocode}
\usepackage{xspace}
\usepackage{float}
\usepackage{multirow}
\usepackage{multicol}
\usepackage{bm}
\usepackage{mathtools}
\usepackage{array}
\usepackage{listings}
\usepackage{ragged2e}

\definecolor{backcolour}{rgb}{0.95,0.95,0.92}
\definecolor{codegreen}{rgb}{0,0.6,0}
\definecolor{codegray}{rgb}{0.5,0.5,0.5}
\definecolor{codepurple}{rgb}{0.58,0,0.82}
\definecolor{ForestGreen}{RGB}{34,139,34}
\definecolor{BrickRed}{RGB}{203,65,84}
\definecolor{verylightblue}{HTML}{f4fbff}
\definecolor{lightblue}{HTML}{77C9FF}
\definecolor{mediumblue}{HTML}{0099FF}
\definecolor{verylightred}{HTML}{fffbf4}
\definecolor{lightred}{HTML}{FF5957}
\definecolor{mediumred}{HTML}{ff002b}
\colorlet{mypink}{red!30}
\colorlet{myblue}{orange!30}
\colorlet{mypurple}{green!10}
\definecolor{customblue}{HTML}{0099ff}
\definecolor{customred}{HTML}{ff006b}

\newcolumntype{L}{>{\RaggedRight\arraybackslash}p{11.5em}}
\newcolumntype{C}{>{\Centering\arraybackslash}X}
\newcolumntype{M}{>{\RaggedRight\arraybackslash}p{11.5em}}

\renewcommand{\today}{January 2026}

\title{Policy of Thoughts: Scaling Test-Time Training for LLM Reasoning via Online Policy Evolution}

\correspondingauthor{kongdezhang@zju.edu.cn, han\_meng@zju.edu.cn}

\author[1*]{Zhengbo Jiao}
\author[1,2*]{Hongyu Xian}
\author[1,3]{Qinglong Wang}
\author[4]{Yunpu Ma}
\author[1]{Zhebo Wang}
\author[5]{Zifan Zhang}
\author[1$\dagger$]{Dezhang Kong}
\author[1$\dagger$]{Meng Han}

\affil[1]{Zhejiang University}
\affil[2]{South China Normal University}
\affil[3]{Shanghai University of Finance and Economics}
\affil[4]{LMU Munich}
\affil[5]{Wuhan University}

\begin{abstract}
Large language models (LLMs) struggle with complex, long-horizon reasoning due to instability from the frozen-policy assumption. Current \textit{test-time scaling} methods treat execution feedback merely as an external signal for filtering or rewriting trajectories, without internalizing it to improve reasoning strategies. We argue that effective reasoning requires \textbf{test-time training (TTT)}---real-time gradient-based policy adaptation through learning from failures. Inspired by Popper's ``conjectures and refutations,'' we introduce \textbf{Policy of Thoughts (PoT)}, a test-time training framework that recasts reasoning as a within-instance online optimization process. PoT performs structured exploration via Monte Carlo Tree Search to generate diverse candidates, then applies Group Relative Policy Optimization (GRPO) to update a transient LoRA adapter using execution feedback---effectively training the model on each instance at test time. The per-instance adapter is discarded after solving, leaving the base model untouched. This closed-loop \emph{test-time training} design enables dynamic, instance-specific refinement of reasoning priors, converting additional compute into direct policy improvement rather than wasted exploration. Experiments demonstrate PoT's substantial gains across multiple benchmarks. On LiveCodeBench, a 4B model achieves 49.71\% accuracy, surpassing 12 test-time scaling baselines and commercial models (e.g., GPT-4o, DeepSeek-V3) despite being over 50$\times$ smaller.
\end{abstract}

\begin{document}

\maketitle

\begin{figure}[htbp]
    \centering
    \includegraphics[width=\linewidth]{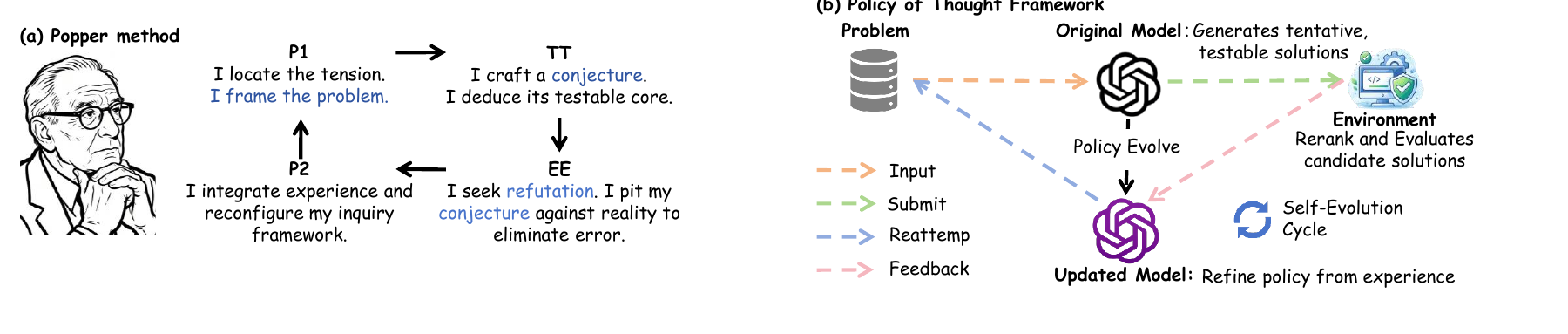}
    \caption{The Policy of Thoughts (PoT) Framework: A Closed-Loop Reasoning System inspired by Popperian epistemology. (a) \textbf{The reasoning cycle:} P1 (identify problem), TT (propose conjecture), EE (test against reality), and P2 (update understanding). (b) \textbf{PoT implementation:} The model generates solutions (TT); the environment evaluates (EE); feedback is internalized via RL to update the reasoning policy (P2), enabling real-time adaptation.}
    \label{fig:pot_framework}
\end{figure}

\section{Introduction}
\label{sec:introduction}
Recent advances are increasingly positioning large language models (LLMs) as general-purpose reasoners, capable of solving problems that require multi-step deduction, intermediate decisions, and long-horizon planning. While eliciting explicit intermediate reasoning steps (Chain-of-Thought) \cite{wei2023chainofthoughtpromptingelicitsreasoning} substantially improves accuracy, reasoning stability remains a central bottleneck on hard instances: these tasks induce a vast search space filled with deceptive trajectories, where a single early mistake can irreversibly derail the entire reasoning chain. We argue that this instability is intrinsic to a frozen policy: without the ability to internalize its own failed attempts, the model lacks a mechanism to consistently correct its course and converge to the correct solution.

The mainstream approach to mitigate this problem is \textit{test-time scaling}---increasing inference-time computation \cite{wang2025surveyparallelreasoning}. Existing methods either expand search width through sampling \cite{wang2023selfconsistencyimproveschainthought, yao2023treeofthoughts, kang2025scalablebestofnselectionlarge} or increase depth through iterative refinement \cite{madaan2023selfrefine, shinn2023reflexionlanguageagentsverbal}, using execution feedback mainly to filter or revise trajectories. However, they still rely on a frozen policy, spending substantial computation on failed attempts without improving the reasoning strategy itself. In AI history, learning has often surpassed search on hard problems, as shown by AlphaGo \cite{silver2017mastering} and AlphaFold \cite{jumper2021highly}. This raises a natural question: \emph{can we learn, rather than merely search, at test time?} Recent test-time training (TTT) methods \cite{akyurek2025ttt, zuo2025ttrl, yuksekgonul2026tttdiscover} explore this direction via gradient-based updates during inference. However, they target cross-instance adaptation or long-horizon scientific discovery rather than the per-instance reasoning setting we study.

\begin{figure}[htbp]
    \centering
    \includegraphics[width=\linewidth]{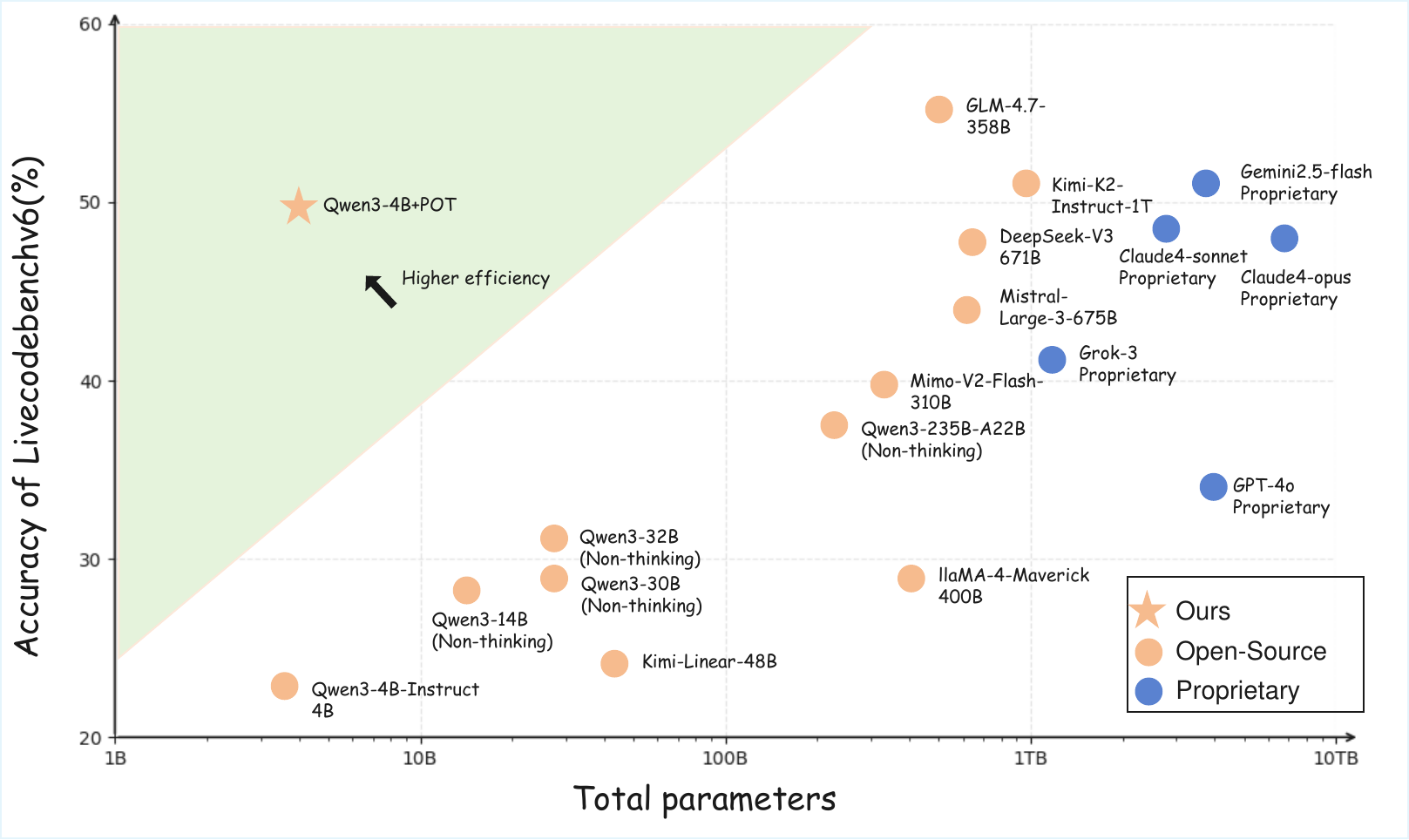}
    \caption{Comparison of model scale versus Live evaluation accuracy. Our PoT-enhanced model achieves 49.71\% accuracy, significantly outperforming larger parameter scales---where accuracy degrades with increasing parameters.}
    \label{fig:accuracy_comparison}
\end{figure}

We introduce \textbf{Policy of Thoughts (PoT)}, a framework that effectively recasts test-time reasoning as a within-instance online optimization process. Inspired by the epistemological principle of ``conjectures and refutations,'' PoT enables the model to iteratively evolve its reasoning policy through a closed loop of exploration and adaptation. Our key design insight is that \emph{per-instance transient adaptation} is the right regime for reasoning problems: unlike standard RL, where the goal is a policy that generalizes across many inputs, each reasoning instance requires its own reasoning strategy, and the adapted policy is merely a means toward finding one correct solution rather than an end in itself \cite{yuksekgonul2026tttdiscover}. Concretely, PoT first explores diverse candidate solutions via Monte Carlo Tree Search (MCTS), then uses Group Relative Policy Optimization (GRPO) \cite{shao2024deepseekmathpushinglimitsmathematical} to update a transient LoRA adapter based on execution feedback. The adapter is discarded after each problem, leaving the base model untouched. This closed-loop design transforms additional test-time compute into direct policy improvement rather than wasted exploration (see Figure~\ref{fig:accuracy_comparison}). Extensive experiments show that PoT with small models consistently outperforms 16 existing test-time scaling methods and five commercial large models across multiple benchmarks.

Our contributions are summarized as follows:
\begin{enumerate}[leftmargin=*,topsep=0.5pt,itemsep=0pt,parsep=0pt]
    \item \textbf{Test-Time Training as Policy Evolution.} We formalize test-time reasoning as a within-instance online optimization (i.e., test-time training) process, shifting the paradigm from static trajectory selection to dynamic policy adaptation. This allows LLMs to transcend the frozen-policy assumption by internalizing execution feedback and reshaping their reasoning strategy in real time.

    \item \textbf{Policy of Thoughts Framework.} We implement the PoT framework, which unifies structured exploration (MCTS) with parameter-efficient internalization (GRPO + transient LoRA). The per-instance adapter design avoids catastrophic forgetting and cross-instance interference, making PoT well-suited to heterogeneous reasoning tasks.

    \item \textbf{Superior Empirical Performance.} A PoT-enhanced 4B model achieves 64.02\% overall accuracy across five code reasoning benchmarks, surpassing GPT-4o (49.35\%), Claude-Opus-4 (55.22\%), and Gemini-2.5-Flash (60.91\%) despite being over 50$\times$ smaller. PoT also improves the reasoning capabilities of thinking models: Qwen3-4B-Thinking accuracy rises from 55.20\% to 63.70\% on LiveCodeBench V6.
\end{enumerate}

\section{Related Work}
\label{sec:related_work}
\textbf{Test-Time Scaling via Sampling and Refinement.}
Chain-of-Thought prompting \citep{wei2023chainofthoughtpromptingelicitsreasoning} established the basis for scaling test-time computation through intermediate reasoning. Subsequent methods such as Self-Consistency \citep{wang2023selfconsistencyimproveschainthought} and Best-of-N \citep{kang2025scalablebestofnselectionlarge} improve accuracy by aggregating sampled trajectories. Iterative refinement further enables self-correction: Self-Refine \citep{madaan2023selfrefine} and Reflexion \citep{shinn2023reflexionlanguageagentsverbal} revise outputs with verbal feedback, while CodeT \citep{chen2022codetcodegenerationgenerated} and Self-Debug \citep{chen2023teachinglargelanguagemodels} use execution feedback to verify and repair code. Recent systems such as OpenAI o1 \citep{openai2024o1systemcard} further combine extended reasoning with self-correction. However, these methods operate under a \emph{frozen policy}: feedback is used only for filtering, rewriting, or prompting, without updating the underlying reasoning strategy.

\textbf{Structured Search and Planning.}
Reasoning can also be formulated as structured search. Tree of Thoughts \citep{yao2023treeofthoughts} and RAP \citep{hao2023reasoning} search over candidate thoughts, later extended by LATS \citep{zhou2024languageagenttreesearch} and planning-based approaches \citep{zhang2023planning}. To improve efficiency, ReST-MCTS \citep{zhang2024restmcts} and process reward-guided ensembling \citep{park2024ensemblinglargelanguagemodels} incorporate finer-grained feedback into Monte Carlo Tree Search. Recent work including AB-MCTS \citep{inoue2025wider} and RethinkMCTS \citep{li2025rethinkmctsrefiningerroneousthoughts} further improves search efficiency and thought refinement. Despite strong exploration ability, these methods still keep the generation policy fixed throughout search.

\textbf{Test-Time Training and Adaptation.}
A growing line of work relaxes the frozen-policy assumption through gradient-based updates at inference. \citet{akyurek2025ttt} first demonstrated test-time training (TTT) for few-shot learning. This idea has since expanded: TTT-E2E \citep{tandon2025ttte2e} applies TTT to long-context modeling, In-Place TTT \citep{cai2026inplacettt} enables adaptation for Transformer LLMs, LaCT \citep{zhang2025lact} targets efficient 3D reconstruction, and Spatial-TTT \citep{liu2026spatialttt} extends TTT to streaming spatial intelligence. In reasoning, TTRL \citep{zuo2025ttrl} performs test-time RL with self-consistency rewards, TTT-Discover \citep{yuksekgonul2026tttdiscover} applies RL-based TTT to scientific discovery, TEMPO \citep{zhang2026tempo} studies scaling through alternating refinement and critic recalibration, and AlphaProof \citep{deepmind2025alphaproof} demonstrates test-time RL for formal reasoning.

\begin{figure*}[htbp] 
    \centering
    \includegraphics[width=\linewidth]{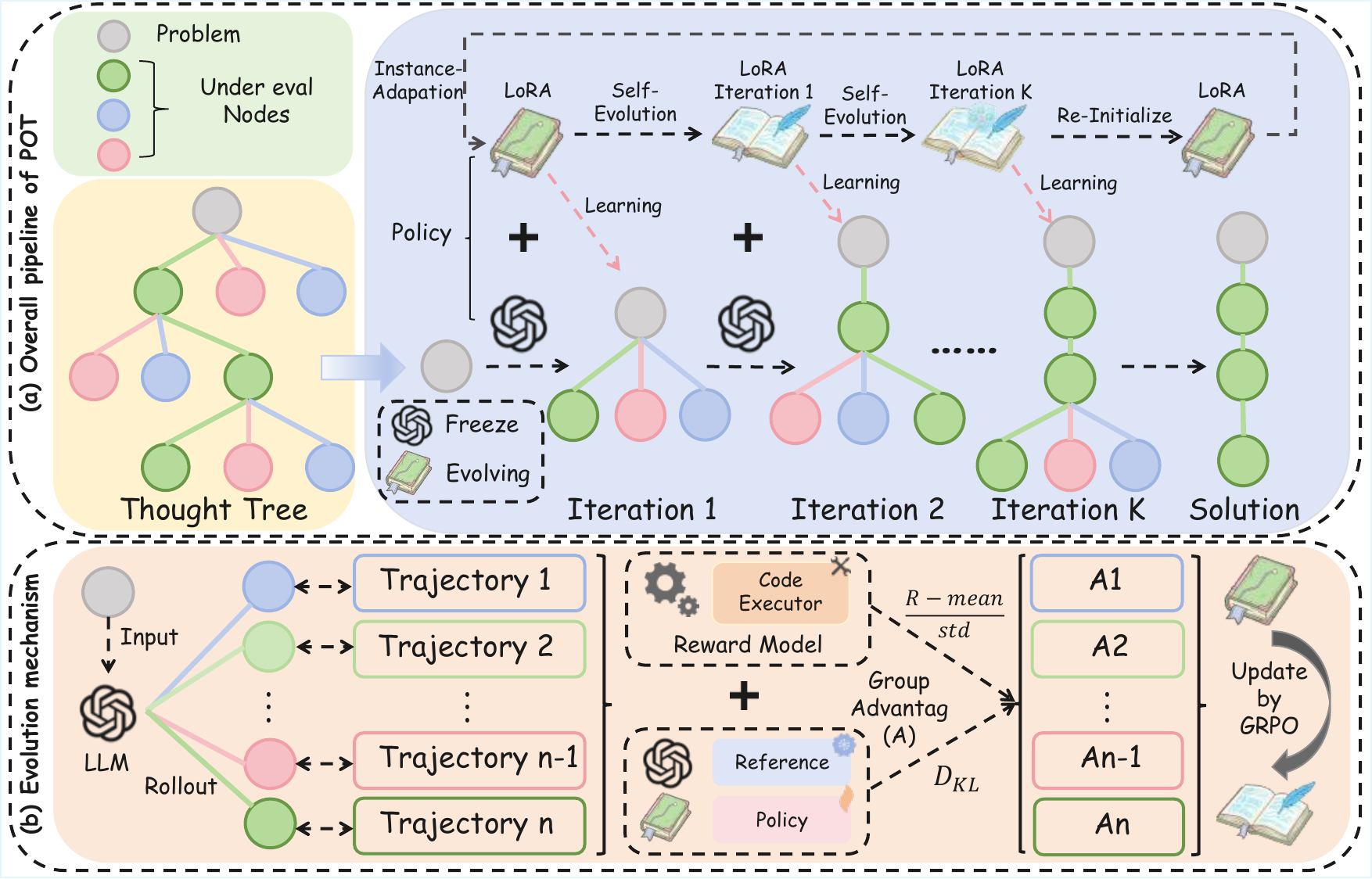}
    \caption{\textbf{Policy evolution on a thought tree with frozen LLM and dynamic LoRA adaptation.} (a) \textbf{Overall pipeline:} A thought tree is constructed rooted at the problem. The LLM backbone remains frozen while the policy evolves through LoRA adapters. In each iteration, children are expanded from pending nodes, evaluated, and low-quality branches are pruned. (b) \textbf{Single iteration mechanism:} All child nodes sharing the same parent form a group. Their trajectories are evaluated and used to compute group-wise advantage signals. The LoRA parameters are then updated via GRPO, enabling structured policy refinement.}
    \label{fig:pot_pipeline}
\end{figure*}

\section{Preliminaries}
\label{sec:preliminaries}

\subsection{Monte Carlo Tree Search}
MCTS is a decision-making framework designed to explore complex solution spaces by incrementally constructing a search tree. In reasoning tasks, each node represents an intermediate thought or a partial solution state. The search process follows four iterative phases: (1) \textbf{Selection}, where nodes are chosen based on the Predictor Upper Confidence Bound applied to Trees (PUCT) to balance exploration and exploitation:
\begin{equation}
a_t = \arg\max_a \left( Q(s,a) + c_{\text{puct}}\,P(s,a) \frac{\sqrt{\sum_{a'} N(s,a')}}{1+N(s,a)} \right).
\label{eq:puct}
\end{equation}
where $Q(s,a)$ is the action value, $N(s,a)$ is the visit count, and $P(s,a)$ is the policy prior. (2) \textbf{Expansion}, where new potential thoughts are generated by the policy; (3) \textbf{Simulation}, where complete reasoning trajectories $\tau$ are sampled to reach a terminal state; and (4) \textbf{Backpropagation}, where the reward $R(\tau)$ from environment feedback is used to update the values of ancestral nodes:
\begin{equation}
Q(s, a) = \frac{1}{N(s, a)} \sum_{i=1}^{N(s, a)} R(\tau_i)
\label{eq:q_update}
\end{equation}
By aggregating terminal outcomes through Equation \ref{eq:q_update}, MCTS filters out erroneous paths and serves as the structured exploration engine for generating diverse candidate trajectories.

\subsection{Low-Rank Adaptation}
To enable parameter-efficient test-time evolution, we adopt \textbf{Low-Rank Adaptation (LoRA)} \citep{lora}. For a frozen pre-trained weight matrix $W_0 \in \mathbb{R}^{d \times k}$, LoRA represents the parameter update $\Delta W$ as the product of two low-rank matrices $B \in \mathbb{R}^{d \times r}$ and $A \in \mathbb{R}^{r \times k}$, where $r \ll \min(d, k)$. The forward pass is modified as:
\begin{equation}
h = W_0 x + \Delta W x = W_0 x + \frac{\alpha}{r} BA x
\label{eq:lora_forward}
\end{equation}
During a reasoning episode, only $A$ and $B$ are subject to optimization. LoRA acts as a \textbf{transient adapter}, providing a lightweight vessel for the rapid parameter mutations needed to transform the initial policy $P_1$ into the evolved state $P_2$.

\subsection{Group Relative Policy Optimization}
We employ GRPO \citep{shao2024deepseekmathpushinglimitsmathematical} to transform external feedback into internal knowledge. For a given problem $q$, GRPO samples a group of $G$ trajectories $\{o_1, o_2, \dots, o_G\}$ and optimizes the policy $\pi_\theta$ by maximizing:
\begin{equation}
\mathcal{L}_{\text{GRPO}}(\theta) = \frac{1}{G} \sum_{i=1}^G \left[ \min \left( \frac{\pi_\theta(o_i|q)}{\pi_{\text{old}}} \hat{A}_i,\; \text{clip} \right) - \beta \, \mathbb{D}_{KL}(\pi_\theta \| \pi_{\text{ref}}) \right]
\label{eq:grpo_loss}
\end{equation}

where the advantage $\hat{A}_i$ is estimated via relative rewards within the sampled group:
\begin{equation}
\hat{A}_i = \frac{r_i - \text{mean}(r_1, \dots, r_G)}{\text{std}(r_1, \dots, r_G)}
\label{eq:grpo_advantage}
\end{equation}
By utilizing group-relative reward normalization, GRPO effectively internalizes environment feedback into the model’s parameters, enabling instance-specific policy adaptation.

\section{Methodology: Policy of Thoughts (PoT)}
\label{sec:method}
In this section, we present the \textbf{Policy of Thoughts (PoT)} framework. We first formalize the reasoning process as an online adaptation problem, followed by an overview of the evolutionary cycle. We then detail the mechanisms of exploratory conjecture and policy (see Figure~\ref{fig:pot_pipeline}).

\subsection{Problem Formulation: Reasoning as Online Adaptation}
We formulate problem solving as a \textbf{finite-horizon Markov Decision Process (MDP)} $(\mathcal{S}, \mathcal{A}, \mathcal{T}, \mathcal{R})$. Unlike static paradigms with a fixed policy, PoT adapts an instance-specific policy $\pi_{\theta,\phi}$ within a single reasoning episode, where $\theta$ denotes frozen pre-trained weights and $\phi$ a \textbf{transient adapter}.


\begin{itemize}
    \item \textbf{State Space $\mathcal{S}$}: A state $s_t = (\mathcal{P}, \mathcal{H}_t)$ consists of the problem description $\mathcal{P}$ and the history $\mathcal{H}_t = \{(c_i, f_i)\}_{i=1}^{t-1}$, where $c_i$ is the $i$-th generated thought and $f_i$ is the corresponding execution feedback.
    \item \textbf{Action Space $\mathcal{A}$}: An action $a_t$ generates a \textbf{Thought} $c_t$, i.e., a functionally complete code implementation as a token sequence $(y_1, \dots, y_L)$, where $y_j$ is the $j$-th token.
    \item \textbf{Transition $\mathcal{T}$}: Transitions are deterministic: \\
    $s_{t+1} = s_t \cup (c_t, f_t)$.
    \item \textbf{Unified Reward $\mathcal{R}$}: Let $\tau = (c_1, c_2, \dots, c_t)$ denote a complete reasoning trajectory—a sequence of thoughts generated for problem $\mathcal{P}$. To quantify the degree of refutation for each conjecture, we define a reward $R(\tau)$ derived from unit test execution:
    \begin{equation}
\small
R(\tau)=
\begin{cases}
1.0, & \text{if all pass (AC)} \\
N_{\text{pass}}/N_{\text{total}}, & \text{if partially pass} \\
0, & \text{otherwise}
\end{cases}
\label{eq:reward_formal}
\end{equation}
    where $N_{\text{pass}}$ and $N_{\text{total}}$ denote the number of passed and total unit tests, respectively.
\end{itemize}

\begin{figure*}[htbp] 
    \centering
    \includegraphics[width=\linewidth]{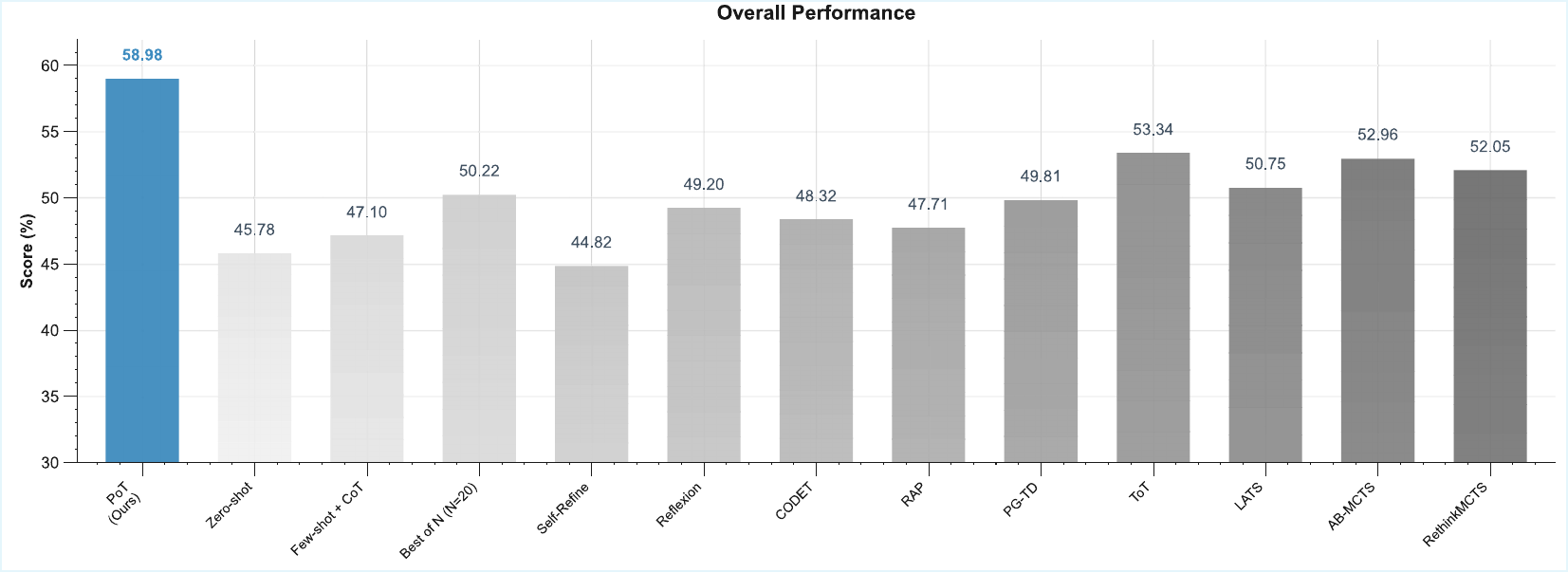}
    \caption{\textbf{Overall performance on reasoning benchmarks.} 
    PoT achieves the highest score (58.98\%), significantly outperforming self-refinement methods, search-based approaches, and standard inference baselines. It also surpasses recent MCTS-enhanced methods such as AB-MCTS and ReinforceMCTS, demonstrating the effectiveness of its unified thought-tree-guided policy evolution framework.}
    \label{fig:placeholder}
\end{figure*}


\subsection{Framework Overview: The Evolution Cycle}
Guided by Popperian epistemology, PoT recasts reasoning as a cycle of \textbf{$P_1 \to TT \to EE \to P_2$} (\textbf{Figure 1}). For any given problem $\mathcal{P}$, PoT initializes a task-specific \textbf{transient adapter} $\phi_1$, representing the initial policy state $P_1$.

The reasoning process unfolds over multiple steps $t \in \{1, \dots, T\}$, each comprising an interleaved phase of exploration and adaptation:
\begin{itemize}
    \item \textbf{Exploratory Conjecture ($TT$)}: PoT employs structured search to explore the solution space, generating a diverse ensemble of trajectories as tentative theories.
    \item \textbf{Policy Internalization ($EE \to P_2$)}: Execution feedback serves as objective refutation. PoT leverages GRPO to \textbf{internalize} these signals into the transient adapter, producing an evolved policy $\phi_{t+1}$ (state $P_2$) with a refined prior for subsequent search.
\end{itemize}
This cycle ensures that test-time computation supports both solution-space exploration and internal reasoning refinement.
\subsection{Exploratory Conjecture via Structured Search}
To generate the diverse conjectures required for policy evolution, PoT performs a search-driven exploration phase. At each step $t$, it initiates $k$ parallel simulations to populate an experience buffer $\mathcal{B}$, where $k$ is a fixed hyperparameter controlling the breadth of exploration.

\textbf{Selection and Expansion.} The search engine traverses the tree by selecting actions that maximize the PUCT objective:
\begin{equation}
a_t = \operatorname*{argmax}_a \left( Q(s,a) + c_{\text{puct}}\, P(s,a) \frac{\sqrt{\sum N(s, \cdot)}}{1+N(s,a)} \right)
\label{eq:puct_method}
\end{equation}
where $N(s, a)$ denotes the visit count, $P(s, a)$ is the prior probability under the current adapter $\phi_t$, and $C > 0$ is the exploration constant. Upon reaching a leaf node, $k$ candidate thoughts are generated ($k$ is a fixed expansion width) and evaluated by environment execution.

\textbf{Backpropagation.} The reward $R(\tau)$ is backpropagated to update the value $\hat{Q}$ along the visited path in the tree.

\textbf{Early Termination.} If any simulation yields a perfect reward ($R(\tau) = 1.0$), the search terminates immediately and returns the corresponding solution. Otherwise, the ensemble of $G$ trajectories $\mathcal{B} = \{ (\tau_i, R(\tau_i)) \}_{i=1}^G$ is forwarded to the internalization phase, where $G$ denotes the group size (i.e., the number of collected trajectories per iteration).

\subsection{Policy Internalization via Online Optimization}
When exploration fails to yield a solution, PoT converts feedback from failed conjectures into policy-level updates, constituting the core of \textbf{test-time policy evolution}.

\textbf{Relative Advantage Estimation.} To internalize refutation signals, we compute the relative advantage $\hat{A}_i$ within the group $\mathcal{B}$. This identifies comparatively more promising conjectures and provides the gradient direction for policy evolution:
\begin{equation}
\hat{A}_i = \text{clip} \!\left( \frac{R(\tau_i) - \mu_G}{\max(\sigma_G, \eta)},\, -C_A,\, C_A \right)
\label{eq:advantage_final}
\end{equation}
where $\mu_G = \text{mean}(\{R_j\}_{j=1}^G)$ and $\sigma_G = \text{std}(\{R_j\}_{j=1}^G)$.
where $\eta > 0$ is a small constant to prevent division by zero, $C_A > 0$ is the clipping bound for advantage normalization, and $\text{mean}(\cdot)$, $\text{std}(\cdot)$ denote the empirical mean and standard deviation over the group.

\textbf{Gradient-based Internalization.} We minimize the GRPO loss to update the transient adapter $\phi$, directly encoding refutation signals into the model's parameters. Let $r_{i,j}(\phi) = {\pi_{\phi}(y_{i,j} \mid s, y_{i,<j})}/{\pi_{\phi_t}(y_{i,j} \mid s, y_{i,<j})}$ denote the importance sampling ratio. The loss is:
\begin{equation}
\begin{split}
\mathcal{L}(\phi) = & \frac{1}{G} \sum_{i=1}^G \frac{1}{|\tau_i|} \sum_{j=1}^{|\tau_i|} \Bigl[
\min \bigl( r_{i,j} \hat{A}_i, \\
& \text{clip}(r_{i,j}, 1{-}\epsilon, 1{+}\epsilon) \hat{A}_i \bigr) \\
& - \beta \, \mathbb{D}_{\mathrm{KL}}(\pi_{\phi} \| \pi_{\text{ref}}) \Bigr]
\end{split}
\label{eq:grpo_final_token}
\end{equation}
Here, $|\tau_i|$ denotes the token count of trajectory $\tau_i$, $y_{i,<j}$ is the token prefix up to position $j{-}1$, $\epsilon > 0$ is the PPO clipping threshold, $\beta \geq 0$ controls KL regularization strength, and $\pi_{\text{ref}}$ is the reference policy (typically $\phi_t$).

\textbf{Action Commitment.} After $E$ epochs of optimization (where $E$ is a fixed number of fine-tuning steps), the model commits to a new thought $c_t$ via greedy decoding from the evolved policy $\pi_{\theta, \phi_{t+1}}$. This updated policy then serves as the prior for the next step’s exploration, progressively narrowing the search toward high-quality regions. Upon episode completion, the transient adapter is discarded, resetting the model for the next independent task.

\section{Experiments}
\label{sec:experiments}

\begin{table*}[tb!]
\centering
\small
\caption{Code reasoning results with a 4B solver. All methods use the same inference budget and evaluation protocol (Appendix~\ref{sec:baseline-setup}). LiveCodeBench (LCB) results are evaluated on the \texttt{code\_generation\_lite} split. Overall is the unweighted average across all five benchmarks.}
\label{tab:main_results_code_math}
\setlength{\tabcolsep}{4pt}
\begin{tabularx}{\textwidth}{@{}L|CCCCC|>{\columncolor{gray!10}}C@{}}
\toprule
Baselines & HumanEval & MBPP & LCB V5 & LCB V6 & ICPC & Overall \\
\midrule

\multicolumn{7}{@{}l}{\textit{\textbf{Model Baselines} (Strong Commercial models (zero-shot))}} \\
$~~~~$Qwen3-235B-A22B & 79.88 & 75.10 & 35.33 & 37.21 & 5.48 & 46.60 \\
$~~~~$DeepSeek-V3 & 91.50 & 87.60 & 31.74 & 16.00 & 9.59 & 47.29 \\
$~~~~$Gemini-2.5-Flash & 94.01 & 81.70 & 64.07 & 56.57 & 8.22 & 60.91 \\
$~~~~$GPT-4o & 92.70 & 87.60 & 29.94 & 29.71 & 6.80 & 49.35 \\
$~~~~$Claude-Opus-4 & 95.80 & 83.20 & 50.30 & 40.00 & 6.80 & 55.22 \\
\midrule

\multicolumn{7}{@{}l}{\textit{\textbf{Method Baselines} (On Qwen3-4B-Instruct-2507)}} \\

\multicolumn{7}{@{}l}{\cellcolor{verylightblue}$~~~~$\textit{Standard Inference}} \\
$~~~~~~~~$+Zero-shot & 84.76 & 79.77 & 30.54 & 27.43 & 6.80 & 45.86 \\
$~~~~~~~~$+Few-shot + CoT & 85.98 & 82.88 & 31.13 & 28.57 & 8.22 & 47.36 \\
$~~~~~~~~$+Best of N (N=20) & 90.85 & 82.10 & 43.61 & 37.71 & 9.59 & 52.77 \\

\multicolumn{7}{@{}l}{\cellcolor{verylightred}$~~~~$\textit{Self-Refinement / Debugging}} \\
$~~~~~~~~$+Self-Refine & 84.15 & 80.13 & 30.54 & 28.00 & 5.48 & 45.66 \\
$~~~~~~~~$+Reflexion & 87.80 & 85.60 & 34.13 & 38.12 & 10.59 & 51.25 \\
$~~~~~~~~$+CODET & 88.41 & 83.79 & 34.43 & 30.13 & 8.22 & 49.00 \\
$~~~~~~~~$+RAP & 87.19 & 81.71 & 31.73 & 28.00 & 8.22 & 47.37 \\

\multicolumn{7}{@{}l}{\cellcolor{blue!3}$~~~~$\textit{Search-Based Reasoning}} \\
$~~~~~~~~$+PG-TD & 89.02 & 83.72 & 38.12 & 34.32 & 10.59 & 51.15 \\
$~~~~~~~~$+ToT & 90.24 & 82.37 & 43.11 & 38.86 & 16.43 & 54.20 \\
$~~~~~~~~$+LATS & 87.80 & 90.66 & 47.51 & 43.43 & 13.69 & 56.62 \\
$~~~~~~~~$+AB-MCTS & 90.85 & 83.66 & 39.52 & 36.00 & 15.07 & 53.02 \\
$~~~~~~~~$+RethinkMCTS & 89.02 & 87.34 & 49.83 & 44.12 & 15.07 & 57.08 \\

\multicolumn{7}{@{}l}{\cellcolor{green!5}$~~~~$\textit{Test-Time Training / Adaptation}} \\
$~~~~~~~~$+TTRL & 91.22 & 86.14 & 48.61 & 41.67 & 13.70 & 56.27 \\
$~~~~~~~~$+TTT-Discover & 93.90 & 88.76 & 51.38 & 44.86 & 15.07 & 58.79 \\
$~~~~~~~~$+RSA & 88.78 & 86.91 & 48.94 & 43.71 & 14.85 & 56.64 \\
$~~~~~~~~$+s1 & 90.37 & 88.02 & 50.11 & 45.03 & 15.66 & 57.84 \\

\rowcolor{red!10}
$~~~~$\textbf{+PoT (Ours)} & \textbf{98.78} & \textbf{94.94} & \textbf{57.49} & \textbf{49.71} & \textbf{19.18} & \textbf{64.02} \\

\bottomrule
\end{tabularx}
\end{table*}

\begin{table*}[tb!]
\centering
\small
\caption{Cross-architecture generalization results. PoT is applied to models of different families and scales, including instruct variants. LCB results use the \texttt{code\_generation\_lite} split. Overall is the unweighted average across evaluated benchmarks.}
\label{tab:small_models_pot}
\setlength{\tabcolsep}{4pt}
\begin{tabularx}{\textwidth}{@{}L|CCCCC|>{\columncolor{gray!10}}C@{}}
\toprule
Method & HumanEval & MBPP & LCB V5 & LCB V6 & ICPC & Overall \\
\midrule

\multicolumn{7}{@{}l}{\textit{\textbf{Phi-4-mini-4k-instruct}}} \\
$~~~~$+zero-shot & 74.39 & 68.09 & 19.90 & 15.42 & -- & 44.45 \\
\rowcolor{red!10}
$~~~~$+PoT (Ours) & \textbf{87.80} & \textbf{82.49} & \textbf{33.53} & \textbf{27.42} & \textbf{--} & \textbf{57.81} \\

\midrule

\multicolumn{7}{@{}l}{\textit{\textbf{Qwen3-1.7B-Thinking}}} \\
$~~~~$+zero-shot & 78.05 & 70.82 & 32.93 & 29.14 & 4.11 & 43.01 \\
\rowcolor{red!10}
$~~~~$+PoT (Ours) & \textbf{90.24} & \textbf{84.05} & \textbf{45.51} & \textbf{42.29} & \textbf{8.23} & \textbf{54.06} \\

\midrule

\multicolumn{7}{@{}l}{\textit{\textbf{Qwen3-30B-A3B-Instruct-2507}}} \\
$~~~~$+zero-shot & 93.30 & 85.20 & 65.00 & 40.20 & 25.97 & 61.93 \\
\rowcolor{red!10}
$~~~~$+PoT (Ours) & \textbf{98.78} & \textbf{96.89} & \textbf{82.04} & \textbf{62.29} & \textbf{41.10} & \textbf{76.22} \\

\bottomrule
\end{tabularx}
\end{table*}

\subsection{Experiment Setup}

\textbf{Models.} We used Qwen3-4B-Instruct-2507 \citep{yang2025qwen3technicalreport}, Qwen3-1.7B-Thinking \citep{yang2025qwen3technicalreport}, Qwen3-30B-A3B-Instruct-2507 \citep{yang2025qwen3technicalreport}, and Phi-4-mini-4k-instruct \citep{phi4mini2025} as the base models on which we deploy PoT and SOTA baseline methods. 

\textbf{Benchmarks.} We evaluated our method on five benchmarks: LiveCodeBench v5 and v6 \citep{jain2024livecodebenchholisticcontaminationfree}, HumanEval \citep{chen2021evaluatinglargelanguagemodels}, MBPP \citep{austin2021programsynthesislargelanguage}, and the ICPC subset of OJBench \citep{wang2025ojbenchcompetitionlevelcode}.The complete benchmark details are provided in Section~\ref{sec:benchmark_details}, which includes the specific number of problems in each dataset: HumanEval (164 problems), MBPP (257 problems), LiveCodeBench v5 lite (167 problems), LiveCodeBench v6 lite (175 problems), and ICPC (73 problems), totaling 836 programming problems spanning various difficulty levels.

\textbf{Baselines.} We evaluate PoT against a comprehensive suite of baselines spanning prompting, iterative refinement, structured search, test-time training, and frontier LLMs. Detailed configurations, hyperparameters, and fairness protocols for all baselines are provided in Appendix~\ref{sec:baseline-setup}.

\textbf{Evaluation Protocol.}
We evaluate all methods using \texttt{pass@1}. To ensure fair comparison, PoT is matched with search-based and TTT baselines under a consistent computational budget; detailed configurations and evaluation protocols are provided in Appendix~\ref{sec:our_setup}.

\textbf{Implementation Details.} See in Appendix \ref{sec:implementation_details}.

\subsection{Main Results}

As shown in Table~\ref{tab:main_results_code_math}, PoT achieves an overall score of 64.02 on Qwen3-4B-Instruct-2507, outperforming the strongest prompting baseline (Best-of-20, 52.77) by +11.25 points, the best search method (RethinkMCTS, 57.08) by +6.94 points, and the best test-time training method (TTT-Discover, 58.79) by +5.23 points. Notably, PoT surpasses commercial models including GPT-4o (49.35), Claude-Opus-4 (55.22), and Gemini-2.5-Flash (60.91) despite being over 50$\times$ smaller. The largest gains appear on challenging benchmarks: +7.66 on LCB V5 and +5.59 on LCB V6, demonstrating that test-time policy evolution is particularly effective for complex reasoning tasks requiring iterative refinement.

\subsection{Cross-Architecture Generalization}

Table~\ref{tab:small_models_pot} shows that PoT generalizes across model families, scales, and training paradigms. On Qwen3-1.7B-Thinking, PoT improves overall accuracy from 43.01 to 54.06 (+11.05 points), while Phi-4-mini-4k-instruct gains +13.36 points. The trend also holds for larger models, with Qwen3-30B-A3B-Instruct-2507 improving by +14.29 points. PoT is also complementary to thinking models; additional results are in Appendix~\ref{sec:thinking_models}. These results confirm that test-time policy evolution is architecture-agnostic and enhances long-form reasoning.

\subsection{Ablation Study}
\label{sec:ablation}
All ablations use Qwen3-4B-Instruct-2507 on LiveCodeBench v6.

\begin{table}[tb!]
\centering
\caption{
Ablation on test-time evolution using Qwen3-4B-Instruct-2507 on LCB v6. 
$\Delta$ denotes absolute percentage point gain relative to the zero-shot baseline.
}
\label{tab:ablation_evolution}
\setlength{\tabcolsep}{4pt}
\renewcommand{\arraystretch}{0.95}
\footnotesize
\begin{tabular}{lcc}
\toprule
\textbf{Configuration} & \textbf{Accuracy (\%)} & \textbf{$\Delta$ (\%)} \\
\midrule
\multicolumn{3}{l}{\textit{Zero-shot}} \\
\quad Base Model & 27.43 & -- \\
\midrule
\multicolumn{3}{l}{\textit{Search Only (No Adaptation)}} \\
\quad \textsc{PoT} w/o LoRA Update & 37.14 & \textcolor{ForestGreen}{$\uparrow$ 9.71} \\
\midrule
\rowcolor{red!10}
\multicolumn{3}{l}{\textit{Search + Adaptation (Ours)}} \\
\rowcolor{red!10}
\quad \textsc{PoT} w/ LoRA Update & \textbf{49.71} & \textbf{\textcolor{ForestGreen}{$\uparrow$ 22.28}} \\
\bottomrule
\end{tabular}

\end{table}

\begin{table}[tb!]
\centering
\caption{
Ablation on branching factor $k$ on LCB v6.
Cost denotes measured wall-clock latency per evolution step.
$\Delta$ is relative to $k=3$ (ours).
}
\label{tab:ablation_k}
\setlength{\tabcolsep}{4pt}
\renewcommand{\arraystretch}{0.95}
\footnotesize
\begin{tabular}{lccc}
\toprule
\textbf{Branching Factor $k$} & \textbf{Accuracy (\%)} & \textbf{$\Delta$ (\%)} & \textbf{Cost (ms)} \\
\midrule
\rowcolor{red!10}
\textbf{$k=3$ (Ours)} & \textbf{49.71} & -- & \textbf{473.66} \\
\midrule
$k=1$ (Greedy) & 31.42 & \textcolor{BrickRed}{$\downarrow$ 18.29} & 161.43 \\
$k=2$ & 41.14 & \textcolor{BrickRed}{$\downarrow$ 8.57} & 309.87 \\
$k=4$ & 50.28 & \textcolor{ForestGreen}{$\uparrow$ 0.57} & 849.47 \\
$k=8$ & 54.86 & \textcolor{ForestGreen}{$\uparrow$ 5.15} & 2369.96 \\
$k=16$ & 55.42 & \textcolor{ForestGreen}{$\uparrow$ 5.71} & 3984.52 \\
$k=32$ & 56.57 & \textcolor{ForestGreen}{$\uparrow$ 6.86} & 8752.96 \\
\bottomrule
\end{tabular}

\end{table}

\begin{table}[tb!]
\centering
\caption{
Ablation on LoRA rank $r$, learning rate $\eta$ on LCB v6.
$\Delta$ denotes absolute difference relative to $(r=8, \eta=10^{-4})$ (ours).
}
\label{tab:ablation_hyper}
\setlength{\tabcolsep}{4pt}
\renewcommand{\arraystretch}{0.95}
\footnotesize
\begin{tabular}{lcc}
\toprule
\textbf{Setting $(r, \eta)$} & \textbf{Accuracy (\%)} & \textbf{$\Delta$ (\%)} \\
\midrule
\rowcolor{red!10}
\textbf{LoRA(8, $1\times10^{-4}$)} (Ours) & \textbf{49.71} & -- \\
\midrule
LoRA(4, $5\times10^{-5}$) & 43.43 & \textcolor{BrickRed}{$\downarrow$ 6.28} \\
LoRA(4, $1\times10^{-4}$) & 46.29 & \textcolor{BrickRed}{$\downarrow$ 3.42} \\
LoRA(8, $5\times10^{-5}$) & 46.86 & \textcolor{BrickRed}{$\downarrow$ 2.85} \\
LoRA(16, $5\times10^{-5}$) & 50.28 & \textcolor{ForestGreen}{$\uparrow$ 0.57} \\
LoRA(16, $1\times10^{-4}$) & 50.86 & \textcolor{ForestGreen}{$\uparrow$ 1.15} \\
\bottomrule
\end{tabular}

\end{table}

\textbf{Ablation on Test-time Evolution.} Table~\ref{tab:ablation_evolution} demonstrates the critical importance of dynamic weight adaptation in \textsc{PoT}. Starting from the Qwen3-4B zero-shot baseline (27.43\%), the version without LoRA updates reaches only 37.14\% through pure MCTS exploration. In contrast, our full framework achieves 49.71\%—a 12.57-point improvement over static search. This margin highlights how on-the-fly internalization of execution feedback enables the model to escape logical plateaus that often limit traditional search, providing essential adaptability for more successful reasoning.

\textbf{Ablation on Branching Factor $k$.} Table~\ref{tab:ablation_k} investigates the impact of search breadth by varying $k$. Results show $k=3$ optimally balances exploration diversity and computational efficiency. Lower values ($k=1,2$) yield significantly worse performance—e.g., $k=1$ drops to 31.42\% due to insufficient contrastive samples for GRPO—while large $k$ incurs prohibitive overhead. Although $k=16$ achieves 55.42\%, its latency (3984.52 ms) is over eight times that of $k=3$ (473.66 ms), indicating severely diminished returns. Thus, $k=3$ provides the ideal equilibrium, delivering robust performance while maintaining high iteration speed for effective test-time evolution.

\textbf{Ablation on LoRA Hyperparameters.} Table~\ref{tab:ablation_hyper} identifies the stability-performance frontier through joint analysis of rank $r$ and learning rate $\eta$. The $(r=8, \eta = 10^{-4})$ configuration achieves the highest accuracy (49.71\%), outperforming others by 2.28--6.28\%. Reducing either $r$ or $\eta$ leads to drops up to 6.28\%, suggesting insufficient capacity for immediate knowledge internalization. Increasing $r$ to 16 yields marginal gains but risks overfitting on sparse per-instance feedback.

\section{Conclusion}
\label{sec:conclusion}
We present Policy of Thoughts (PoT), a test-time reasoning framework that moves beyond the frozen-policy assumption by enabling per-instance online policy evolution. PoT integrates structured exploration (MCTS) with gradient-based internalization (GRPO + transient LoRA), allowing the model to dynamically refine its reasoning strategy by learning from failed attempts rather than discarding them. Across five code reasoning benchmarks, a PoT-enhanced 4B model achieves 64.02\% overall accuracy---surpassing 16 test-time scaling baselines including recent TTT methods (TTT-Discover, TTRL), as well as commercial models such as GPT-4o (49.35\%), Claude-Opus-4 (55.22\%), and Gemini-2.5-Flash (60.91\%), despite being over 50$\times$ smaller. PoT is also complementary to thinking models, further boosting Qwen3-4B-Thinking from 55.20\% to 63.70\% on LiveCodeBench V6. Ablations confirm that the core gain stems from coupling adaptation with search: pure exploration yields only marginal improvement, while the per-instance transient adapter design outperforms persistent cross-instance adaptation by 5.41 points.

\section*{Limitations}

We acknowledge that the reliance on learned reward models as proxies for ground-truth feedback introduces a degree of subjectivity. Whether model-based signals can fully substitute for deterministic environmental feedback remains an open question and represents an interesting direction for future investigation.

\bibliographystyle{abbrvnat}
\bibliography{references}

\clearpage
\appendix

\section{Generalization and Future Directions}
\label{sec:generalization}

To assess the generality of PoT beyond code generation, we evaluate it across diverse reasoning domains spanning theorem proving, mathematical reasoning, knowledge-intensive QA, and multimodal understanding.

A clear trend emerges across these tasks. PoT exhibits the strongest gains in settings with \textit{native environment feedback}, such as formal proof checkers or tool-execution environments, where refutation signals are precise, deterministic, and directly actionable. In contrast, for tasks without executable interfaces (e.g., open-ended math or knowledge QA), we employ RRM-7B as a unified verifier to score candidate solutions. While performance in these settings is naturally bounded by verifier reliability, PoT still achieves consistent improvements.

We evaluate PoT across five representative benchmarks using reported baseline scores and PoT-enhanced results. As shown in Table~\ref{tab:generalization}, the largest improvements occur in PutnamBench and DeepResearch—both of which provide full environment feedback—whereas RM-based tasks show more moderate but stable gains.

These results demonstrate PoT's broad effectiveness across heterogeneous reasoning tasks, while highlighting reliable reasoning verifiers as a promising future direction.

\begin{table*}[h]
\centering
\caption{Generalization of PoT across reasoning domains. Baseline scores are from official reports. All models use non-thinking mode unless otherwise noted. Tasks with native environment feedback (PutnamBench, DeepResearch) show substantially larger gains compared to RM-guided tasks, though RRM-7B still provides effective signals.}
\label{tab:generalization}
\resizebox{\linewidth}{!}{
\begin{tabular}{l l l c c l}
\toprule
Domain & Benchmark & Model & Baseline & +PoT & Feedback Type \\
\midrule
STEM + General Knowledge & MMLU-Pro & Qwen3-8B & 63.4 & 74.1 & RM (RRM-7B) \\
Non-formal Math Reasoning & AIME-25 & Qwen3-8B & 20.9 & 32.3 & RM (RRM-7B) \\
Formal Math Proof & PutnamBench & DeepSeek-Prover-V2-7B & 9.0 & 30.2 & Native Environment \\
Multimodal Math Reasoning & MathVista & Qwen3-VL-8B & 53.9 & 65.4 & RM (RRM-7B) \\
DeepResearch & TriviaQA & Qwen2.5-7B-Instruct & 59.7 & 75.0 & Native Environment \\
\bottomrule
\end{tabular}
}
\end{table*}

\paragraph{Ablation on Reward Signal Quality.}
To disentangle the impact of verifier capability on PoT's performance in non-executable domains, we conduct an ablation study on AIME-25 using different reward sources (Table~\ref{tab:ablation_rm}). We compare the standard setup against \textit{Self-as-RM} (where the policy model evaluates its own outputs) and stronger proprietary or open-source reasoning RMs (RRM-32B, Gemini-3-Pro, GPT-5).

\begin{table*}[h]
\centering
\caption{Ablation study on AIME-25: Impact of Reward Model choice on PoT performance. The backbone is Qwen3-8B in non-thinking mode. ``Self-as-RM'' denotes the backbone scoring its own solutions. Stronger reasoning RMs yield gains closer to those seen in environment-grounded tasks.}
\label{tab:ablation_rm}
\begin{tabular}{l c c c}
\toprule
Reward Model Source & Type & Accuracy (\%) & $\Delta$ vs Baseline \\
\midrule
\textit{Baseline (No PoT, Qwen3-8B)} & - & 20.9 & - \\
\midrule
Self-as-RM (Qwen3-8B) & Policy Self-Check & 28.4 & +6.5 \\
RRM-7B & Open Weights & 32.3 & +11.4 \\
RRM-32B & Open Weights & 35.1 & +14.2 \\
Gemini-3-Pro & Proprietary API & 35.8 & +14.9 \\
GPT-5 & Proprietary API & 36.2 & +15.3 \\
\bottomrule
\end{tabular}
\end{table*}

The results reveal a clear hierarchy: Reasoning RMs are effective surrogates: Specialized verifiers like RRM-7B/32B bridge the gap significantly, and scaling to frontier models (Gemini-3-Pro, GPT-5) further amplifies performance. This indicates that while \textit{environment-dense} tasks (like coding) naturally favor PoT due to deterministic feedback, the framework is robust enough to leverage high-quality \textit{reasoning-based} feedback to achieve comparable trajectories in abstract domains.

These findings underscore a key insight: PoT is best suited for \textit{agentic, grounded reasoning}. However, as reward models evolve from simple classifiers to complex reasoners, the distinction between "environment feedback" and "RM feedback" begins to blur, allowing PoT to extend its efficacy into purely conceptual domains. Future work will focus on narrowing the gap between open-weight RMs and ground-truth environments for scientific discovery.

\section{Baseline Setup}
\label{sec:baseline-setup}

We unify the execution environment for all baselines to ensure consistent comparison. All baselines run under identical inference hardware and software configurations, with shared backend, timeout settings, and dataset processing pipelines.

\paragraph{Simple (0-shot).}
Temperature is set to $0.0$ with greedy decoding. A single code solution is generated without iterative refinement. The maximum output length per generation is set to $8192$ tokens. For MBPP, a 3-shot prompt template is applied. This baseline is evaluated on HumanEval, MBPP, LiveCodeBench, and ICPC.

\paragraph{Few-shot + CoT.}\citep{wei2023chainofthoughtpromptingelicitsreasoning}
This baseline provides the model with few-shot exemplars that demonstrate intermediate reasoning steps in a chain-of-thought style. The prompt includes task-specific exemplars, input-output formats, and target problem instructions. Decoding uses temperature $0.0$ to enforce deterministic reasoning. The token budget for generation is set to $16,384$ tokens to ensure sufficient space for long-form reasoning chains. No iterative refinement or feedback mechanism is applied; only a single forward generation is executed per task.

\paragraph{Best-of-N.}\citep{kang2025scalablebestofnselectionlarge}
This baseline samples $N=20$ candidates using temperature $0.7$. Each candidate is generated with a maximum length of $8192$ tokens. If any candidate passes all test cases, execution stops early. This method emphasizes sampling-based diversity.

\paragraph{Reflexion.}\citep{shinn2023reflexionlanguageagentsverbal}
This baseline performs iterative refinement with at most $10$ refinement rounds. Each round conducts $2$ internal test executions, then the model reflects and regenerates using temperature $0.7$. The maximum output length per generation is $8192$ tokens. The total number of LLM calls is approximately $20$ (generation + reflection).

\paragraph{Self-Refine.}\citep{madaan2023selfrefine}
This baseline executes a single initial generation, followed by $3$ refinement rounds. Generation uses temperature $0.7$, feedback uses temperature $0.3$. The maximum output length per generation is $2048$ tokens. No code execution is performed; the model internally assesses correctness and may stop early.

\paragraph{ToT (Tree of Thoughts).}\citep{yao2023treeofthoughts}
This baseline performs tree-structured search with maximum depth $5$, beam width $3$, and $3$ sampled candidates per node. Sampling uses temperature $0.7$, and the maximum output length per node is $2048$ tokens. Search proceeds via BFS with scoring and pruning.

\paragraph{LDB (Large Language Model Debugger).}\citep{zhong2024debug}
This baseline segments a program into basic blocks, executes it on a visible test case to collect runtime variable states, and queries the LLM once per debugging iteration with all selected block-level execution traces; the prompt is capped at $3,097$ input tokens, at most $10$ basic blocks are included per iteration, greedy decoding (temperature $0$) is used, and up to 10 debugging iterations are allowed.

\paragraph{RAP (Reasoning via Planning).}\citep{hao2023reasoning}
This baseline uses Monte Carlo Tree Search (MCTS) with 20 iterations, sampling exactly $3$ actions per expansion, a maximum reasoning depth of $8$, and each LLM call (for action or state generation) capped at $2048$ tokens with temperature $0.8$.

\paragraph{PG-TD (Planning-Guided Transformer Decoding).}\citep{zhang2023planning}
This baseline performs tree search over LLM-generated code solutions, where each node represents a partial program; at each leaf, it uses beam search ($beam size = 1$) to complete the program and evaluates it on public test cases to assign a pass-rate reward; the search uses P-UCB with exploration weight $c=4$, expands the top-3 most likely next tokens per node, and runs for up to $256$ total LLM calls (one per rollout); each LLM generation is capped at $1024$ tokens and uses temperature $0.7$.

\paragraph{LATS (Language Agent Tree Search).}\citep{zhou2024languageagenttreesearch}
This baseline performs at most $10$ search iterations, each expanding $3$ nodes. Each node is tested using $2$ internal executions. Sampling uses temperature $0.7$, with a maximum output length of $4096$ tokens. Node selection follows UCB1 with exploration weight $1.414$, and the search loop consists of selection, expansion, evaluation, and backtracking.

\paragraph{AB-MCTS (Adaptive Branching Monte Carlo Tree Search).}\citep{inoue2025wider}
This baseline performs tree search over LLM-generated code solutions, where each node represents a candidate program evaluated by execution on visible test cases; it uses temperature $0.7$, allows up to $128$ total LLM calls per problem, and dynamically decides at each node whether to generate a new candidate or refine an existing one based on Bayesian posterior updates of observed scores; the method has no fixed branching factor and caps each LLM generation at $1024$ tokens.

\paragraph{RethinkMCTS.}\citep{li2025rethinkmctsrefiningerroneousthoughts}
This baseline applies MCTS with $16$ rollouts, $3$ expansions per rollout, and up to $2$ rethink attempts per failed branch. Sampling uses temperature $0.7$. The maximum decoding length per expansion is $4096$ tokens. A variable P-UCT policy is used with constants $4.0$ and $10.0$.

\paragraph{CodeT.}\citep{chen2022codetcodegenerationgenerated}
This baseline generates $10$ candidate solutions and synthesizes $5$ test suites with $5$ test cases each. Sampling uses temperature $0.8$, and the maximum output length per candidate is $16384$ tokens. Candidate ranking uses Dual Agreement based on test execution outcomes.

\paragraph{Fairness Policy.}
All baselines are executed under the same hardware, inference backend, timeout configuration, and dataset handling pipeline. Token generation limits, sampling configurations, and evaluation methodologies are fully disclosed to ensure transparent and reproducible comparison.

\paragraph{Our Setting.}
Our method adopts a structured search paradigm with explicit tree expansions. At most $M=20$ expansions are allowed, each expansion generates up to $k=3$ child candidates, and each candidate receives up to $2048$ tokens. This yields a maximum search capacity of $60$ generated nodes under the same infrastructure as baselines.

\paragraph{Response Length and Budget Fairness.}
A natural question is whether the different per-candidate token limits across methods (e.g., 8192 for Best-of-N vs.\ 2048 per node for PoT) constitute an unfair comparison. We argue that fairness should be defined at the level of \textit{total computational budget} rather than per-call token length, for the following reasons.

First, different methods have fundamentally different inference structures. Best-of-N requires each candidate to be a \textit{standalone, complete} solution generated from scratch, so it needs a large per-candidate budget. In contrast, PoT operates as a multi-round search-and-adapt process: each node builds upon previously explored context, and the MCTS tree structure allows the model to continue refining a solution across iterations rather than regenerating it from scratch each time. This is analogous to how a human programmer debugs iteratively---each attempt does not restart from zero but builds on prior understanding. The per-node token limit reflects this iterative nature: a 2048-token budget per node is sufficient because the model's adapter has already internalized feedback from earlier attempts.

Second, controlling total budget is strictly more meaningful. Under our protocol, PoT's total token budget is $20 \times 3 \times 2048 = 122{,}880$ tokens, while Best-of-N uses $20 \times 8{,}192 = 163{,}840$ tokens---\textit{a 33\% larger budget for Best-of-N}. Enforcing identical per-call limits would either inflate PoT's total budget (favoring PoT) or compress Best-of-N's per-candidate budget (favoring PoT even more), neither of which constitutes a fair comparison.

Third, we verify empirically that per-candidate length is not the source of PoT's advantage. We ran a unified 2048-token cap across all methods on LCB V6:

\begin{table}[h]
\centering
\small
\caption{Unified 2048-token cap experiment on LCB V6. Capping all methods to the same per-candidate token limit produces minimal changes and preserves the original ranking.}
\label{tab:unified_length}
\begin{tabular}{lcc}
\toprule
\textbf{Method} & \textbf{Original (\%)} & \textbf{Unified 2048-cap (\%)} \\
\midrule
Best-of-N & 37.71 & 36.86 \\
AB-MCTS & 36.00 & 36.27 \\
RethinkMCTS & 44.12 & 43.57 \\
\rowcolor{red!10}
\textbf{PoT (Ours)} & \textbf{49.71} & \textbf{48.94} \\
\bottomrule
\end{tabular}
\end{table}

The ranking is unchanged, and all methods experience comparably small drops ($<$1 point). This confirms that PoT's gains arise from its policy evolution mechanism, not from differences in per-candidate token budgets.

Finally, our compute efficiency analysis (Table~\ref{tab:compute_efficiency}) provides a more comprehensive fairness measure: PoT achieves the best accuracy while consuming only 3.86 PFLOPs per problem, which is \textit{lower} than most search-based and TTT baselines (5--7 PFLOPs). This demonstrates that PoT is not only fair but actually more compute-efficient than the methods it outperforms.

\section{Robustness Analysis}

\begin{table}[tb!]
\centering
\caption{Multi-seed evaluation on LCB V6 (8 independent runs with different random seeds).}
\label{tab:multi_seed}
\setlength{\tabcolsep}{4pt}
\renewcommand{\arraystretch}{0.95}
\footnotesize
\begin{tabular}{lc}
\toprule
\textbf{Method} & \textbf{8-run Mean $\pm$ Std. (\%)} \\
\midrule
Best-of-N & 37.64 $\pm$ 0.61 \\
RethinkMCTS & 44.09 $\pm$ 0.47 \\
\rowcolor{red!10}
\textbf{PoT (Ours)} & \textbf{49.76 $\pm$ 0.34} \\
\bottomrule
\end{tabular}
\end{table}

\begin{table}[tb!]
\centering
\caption{Degenerate GRPO group analysis on LCB V6. A group is degenerate when all $k{=}3$ candidates receive identical rewards, producing zero gradient signal.}
\label{tab:degenerate_group}
\setlength{\tabcolsep}{4pt}
\renewcommand{\arraystretch}{0.95}
\footnotesize
\begin{tabular}{lcc}
\toprule
\textbf{Temperature} & \textbf{Degen. Ratio (\%)} & \textbf{Acc. (\%)} \\
\midrule
0.2 & 16.8 & 56.8 \\
0.4 & 14.2 & 57.3 \\
0.6 & 12.9 & 57.7 \\
0.8 & 15.1 & 57.0 \\
\bottomrule
\end{tabular}
\end{table}

\textbf{Multi-seed Stability.} Table~\ref{tab:multi_seed} reports 8-run mean and standard deviation on LCB V6. PoT achieves the highest mean accuracy (49.76\%) with the lowest variance ($\pm$0.34), confirming that the performance gains are robust and not artifacts of seed selection.

\textbf{Degenerate Group Analysis.} Table~\ref{tab:degenerate_group} examines how often all candidates in a GRPO group receive identical rewards (producing no gradient signal). The degenerate ratio remains moderate (12.9--16.8\%) across temperatures, and accuracy stays stable. Importantly, even when a degenerate group occurs, subsequent exploration still conditions on accumulated context, so the overall search-and-adaptation process does not stall.

\textbf{Ablation on LoRA Hyperparameters.} Table~\ref{tab:ablation_hyper} identifies the stability-performance frontier through joint analysis of rank $r$ and learning rate $\eta$. The $(r=8, \eta = 10^{-4})$ configuration achieves the highest accuracy (49.71\%), outperforming others by 2.28–6.28\%. Reducing either $r$ or $\eta$ leads to drops up to 6.28\%, suggesting insufficient capacity for immediate knowledge internalization. Increasing $r$ to 16 yields marginal gains but risks overfitting on sparse per-instance feedback. Notably, our approach exhibits a form of \textit{benign overfitting}: despite performing intensive, instance-specific updates using only limited execution signals, the model avoids catastrophic forgetting or policy collapse. Instead, it significantly improves task performance while preserving its base reasoning capabilities. This indicates that PoT’s transient adaptation mechanism effectively channels overfitting toward productive refinement rather than degradation.
\textbf{Per-Instance Reset vs.\ Persistent Adapter.} A natural question is whether the transient LoRA adapter should be discarded after each problem or retained across instances. We evaluate a \textit{no-reset} variant on LCB V6 where the adapter is carried over between consecutive problems. This variant achieves 44.3\%, compared with 49.71\% for the standard per-instance reset, a drop of 5.41 points. The degradation suggests that instance-specific updates are not reliably transferable across problems and can introduce cross-instance interference. This validates our design choice of per-instance transient adaptation: the adapted policy is a means toward solving the current problem, not a persistent artifact.

\section{Computational Budget of POT}
\label{sec:our_setup}

To ensure reproducibility and transparency regarding the computational footprint of our method, we explicitly define the resource constraints for the inference search process and the on-the-fly training updates used in POT.

\subsection{Evaluation Protocol and Fairness Constraints}
\label{sec:eval_protocol}

We adopt \texttt{pass@1} as the primary evaluation metric across all experiments.

To ensure rigorous comparison, we align PoT with search-based and TTT baselines under a consistent computational budget by enforcing the same node expansion limit.

For code generation benchmarks, the test cases used for reward computation during PoT's search and adaptation are strictly disjoint from the hidden test cases used for final \texttt{pass@1} evaluation. The model never accesses hidden evaluation tests during inference, search, or parameter updates.

Additional implementation details and hyperparameter settings are provided below.

\subsection{Computational Budget of POT}
\label{sec:compute_budget}
\paragraph{Inference Budget.}
For each problem instance, we set a maximum budget of $M=20$ MCTS iterations. In every iteration, the tree expands by $k=3$ child nodes, resulting in a theoretical maximum of $60$ generated nodes per problem. Each node generation is constrained by a local token limit of $L_{node}=2048$. Although this search configuration defines a high upper bound, the average path depth in practice is effectively regulated by the objective-driven nature of the MCTS selection policy.

\paragraph{On-the-fly Training Overhead.}
The core mechanism of POT involves integrating online GRPO updates directly during the search. We utilize LoRA (rank $r=8$) to update the policy weights efficiently. As demonstrated in our empirical profiling in Table~\ref{tab:overhead_ratio}, the computational cost is manageable. For a batch of $k=3$ nodes (at a representative sequence length), a single LoRA backward pass incurs approximately \textbf{1.46$\times$} the latency of the corresponding forward pass. The total budget approximation is given by:
\begin{equation}
\label{eq:budget_formula}
\begin{split}
\text{Total Budget} 
    &\approx M \times \bigl(\text{Cost}_{\text{fwd}, k} + \text{Cost}_{\text{bwd}}\bigr) \\
    &\approx M \times (1 + 1.46) \times \text{Cost}_{\text{fwd}, k=3}
\end{split}
\end{equation}
This relationship indicates that the additional overhead for online training is roughly equivalent to adding about 1.5 extra node expansion steps (of width $k=3$) per iteration.

\paragraph{Efficiency via Fast Convergence: A Case Study.}
Although the backward pass increases the per-step cost, POT significantly reduces the total number of iterations required to reach the optimal solution. Consider \textit{Jump Game II}, where the model must overcome a strong prior for $O(N^2)$ dynamic programming in favor of an $O(N)$ greedy approach:
\begin{itemize}
    \item \textbf{Static Baseline Scenario:} Without online adaptation, the search policy often remains trapped in local optima, requiring an extensive budget (e.g., $M_{static} \approx 104$) to stochastically sample the correct path using only forward expansions.
    \item \textbf{POT Scenario:} By performing gradient updates on failed candidates, POT suppresses sub-optimal reasoning and converges within a much stricter budget, typically requiring far fewer steps (e.g., $M_{POT} \approx 15$).
\end{itemize}

The computational efficiency gain is calculated by comparing the total cost of static iterations against the cost-augmented POT iterations:
\begin{equation}
\begin{split}
\frac{\text{Cost}_{\text{Static}}}{\text{Cost}_{\text{POT}}} 
    &\approx \frac{104 \times \text{Cost}_{\text{fwd}, k=3}}{15 \times \text{Cost}_{\text{total}, k=3}} \\
    &= \frac{104 \times 192.66}{15 \times 473.66} \approx \mathbf{2.82\times}
\end{split}
\end{equation}
This demonstrates that despite the training overhead, POT achieves superior end-to-step efficiency by preventing wasted exploration in suboptimal reasoning branches.

\begin{table*}[h]
\centering
\caption{Profiling of computational cost components for POT ($k=3$). Total cost reflects the real-time latency as defined in Table~\ref{tab:ablation_k}, where $\text{Cost}_{total} = \text{Cost}_{fwd} + \text{Cost}_{bwd}$.}
\label{tab:overhead_ratio}
\begin{tabular}{lc}
\toprule
\textbf{Component (Batch Size $k=3$)} & \textbf{Computational Cost (ms)} \\
\midrule
Forward Pass ($\text{Cost}_{fwd}$) & 192.66 \\
LoRA Backward ($\text{Cost}_{bwd}$, Rank $r=8$) & 281.00 \\
\midrule
\textbf{Total Iteration Cost} & \textbf{473.66} \\
\textbf{Overhead Ratio ($\text{Cost}_{bwd} / \text{Cost}_{fwd}$)} & \textbf{1.46$\times$} \\
\bottomrule
\end{tabular}
\end{table*}

\paragraph{Global Efficiency vs. Performance.}
As demonstrated in Table~\ref{tab:efficiency_comparison}, POT achieves a state-of-the-art accuracy of 98.78\%, significantly outperforming all baselines. Crucially, our reported time (196.42s) is a total end-to-end duration—strictly encompassing both the LoRA-based backward updates and the MCTS search process—whereas baseline figures (e.g., LATS and AB-MCTS) account for inference only. 

Despite including the full training overhead, POT maintains a total latency comparable to, or even lower than, several leading search-oriented approaches like RethinkMCTS (223.27s). This results in a superior efficiency-performance trade-off: by leveraging gradient-driven adaptation to prune search spaces and suppress sub-optimal reasoning early, POT successfully transmutes backward computation into high-precision reasoning. Compared to static search methods that often plateau around 90\% accuracy even with extensive budgets, POT demonstrates that on-the-fly learning is a more effective path toward solving complex reasoning tasks without incurring prohibitive total time costs.

\begin{table}[t]
\centering
\small
\setlength{\tabcolsep}{4pt}

\caption{
Efficiency and performance comparison on HumanEval using
Qwen3-4B-Instruct-2507.
For PoT, Avg.\ Time includes both LoRA training and inference;
for other baselines, inference only.
}
\label{tab:efficiency_comparison}

\begin{tabular}{
l
>{\centering\arraybackslash}p{1.35cm}
>{\centering\arraybackslash}p{1.45cm}
>{\centering\arraybackslash}p{1.25cm}
}
\toprule

\textbf{Method}
&
\textbf{Training Req.}
&
\textbf{Avg.\ Time (s)}
&
\textbf{Accuracy (\%)}
\\

\midrule

\rowcolor{gray!5}
\multicolumn{4}{l}{\textit{Standard Prompting \& Ensemble}}
\\

Zero-shot   & No & 28.51  & 84.76 \\
Few-shot CoT & No & 52.13  & 85.98 \\
Best-of-N   & No & 113.87 & 90.85 \\

\midrule

\rowcolor{gray!5}
\multicolumn{4}{l}{\textit{Iterative Refinement}}
\\

Self-Refine & No & 58.96 & 84.15 \\
Reflexion   & No & 61.29 & 87.80 \\
CodeT       & No & 56.34 & 88.41 \\
LDB         & No & 71.23 & 87.19 \\

\midrule

\rowcolor{gray!5}
\multicolumn{4}{l}{\textit{Search-based Approaches}}
\\

RAP         & No & 162.33 & 89.63 \\
PG-TD       & No & 178.26 & 89.02 \\
ToT         & No & 121.46 & 90.24 \\
LATS        & No & 178.83 & 87.80 \\
AB-MCTS     & No & 189.24 & 90.85 \\
RethinkMCTS & No & 223.27 & 89.02 \\

\midrule

\rowcolor{red!10}
\textbf{PoT (Ours)}
&
\textbf{Yes (LoRA)}
&
\textbf{196.42}
&
\textbf{98.78}
\\

\bottomrule
\end{tabular}
\end{table}

\section{Implementation Details}
\label{sec:implementation_details}

In this section, we detail the experimental configurations and hyperparameter settings for the PoT framework. Our system consists of a tree-based search mechanism for test-time scaling and a group-relative policy optimization (GRPO) framework for alignment.

\subsection{MCTS Search and Code Verification}
To enhance the model's reasoning trajectory exploration, we implement a Monte Carlo Tree Search (MCTS) strategy. The search process is governed by the Upper Confidence Bound (UCB) formula with an exploration constant $c_{puct} = 1.414$. 

For each simulation, the model generates $k=3$ candidate continuations (generation batch size) with a maximum search depth of $8$. The generation process is constrained to a maximum of $2048$ new tokens per node with a sampling temperature of $T=0.7$. To ensure the technical correctness of the synthesized code, we integrate the LiveCodeBench environment for real-time execution-based verification. We set a code execution timeout of $10$ seconds and assign a reward weight of $1.0$ for successful test case completion. Furthermore, the system incorporates an error feedback mechanism, where execution errors are utilized to refine the search process. The detailed hyperparameters are summarized in Table~\ref{tab:mcts-hyperparams}.

\begin{table}[h]
\centering
\small
\caption{Hyperparameters for MCTS Search and Code Verification.}
\label{tab:mcts-hyperparams}
\begin{tabular}{llc}
\toprule
\textbf{Component} & \textbf{Parameter} & \textbf{Value} \\
\midrule
& Number of simulations & 20 \\
& Temperature ($T$) & 0.7 \\
MCTS Core & Maximum search depth & 8 \\
& Generation batch size ($k$) & 3 \\
& Max new tokens & 2048 \\
\midrule
& Code verification & Enabled \\
Code Evaluation & Code reward weight & 1.0 \\
& Code timeout (seconds) & 10 \\
& Error feedback & Enabled \\
\midrule
Selection & Exploration constant ($c_{puct}$) & 1.414 \\
\bottomrule
\end{tabular}
\end{table}

\subsection{Training and GRPO Parameters}
The model is fine-tuned using Group Relative Policy Optimization (GRPO). For each input prompt, we sample a group of $G=3$ outputs to estimate the advantage function through relative reward ranking, thereby eliminating the need for a separate value network.

The training objective incorporates a KL divergence penalty with a coefficient $\beta = 0.02$ to ensure policy stability. Additionally, a fixed KL coefficient of $0.005$ is applied as a baseline constraint. To prevent excessive drift from the initial policy, the reference model is synchronized every $10$ update steps. We adopt Parameter-Efficient Fine-Tuning (PEFT) via Low-Rank Adaptation (LoRA) with a rank $r=8$ and a corresponding vLLM backend configuration. The optimization is conducted using a learning rate of $1 \times 10^{-4}$ with a PPO clip ratio $\epsilon = 0.3$ over $3$ epochs. The complete training hyperparameters are provided in Table~\ref{tab:training-hyperparams}.

\begin{table}[h]
\centering
\small
\caption{Hyperparameters for GRPO Training and LoRA.}
\label{tab:training-hyperparams}
\begin{tabular}{llc}
\toprule
\textbf{Component} & \textbf{Parameter} & \textbf{Value} \\
\midrule
& Learning rate & $1 \times 10^{-4}$ \\
Optimization & PPO epochs & 3 \\
& PPO clip ratio ($\epsilon$) & 0.3 \\
\midrule
& LoRA rank ($r$) & 8 \\
LoRA (PEFT) & vLLM max LoRA rank & 8 \\
\midrule
& Group size ($G$) & 3 \\
& KL penalty coefficient ($\beta$) & 0.02 \\
GRPO & Fixed KL coefficient & 0.005 \\
& KL ref update frequency & 10 \\
\bottomrule
\end{tabular}
\end{table}

\subsection{Computing Infrastructure}
\label{appendix:hardware}

To ensure the reproducibility of our results, we detail the hardware and software environment used for both training and inference. All experiments were conducted on a single-node workstation equipped with a single NVIDIA A800 (80GB) GPU. The generous $80$GB of HBM2e memory allows for efficient execution of Group Relative Policy Optimization (GRPO) and accommodates the significant KV cache demands during deep MCTS simulations.

The software stack is built upon the latest industry standards to leverage high-performance kernels. We utilize CUDA 13.0 and PyTorch 2.9.0 as the primary deep learning framework. For the test-time scaling phase, we employ vLLM (v0.13.0) as the inference engine, which provides optimized support for the LoRA adapter and high-throughput continuous batching during tree search.

\begin{table}[h]
\centering
\small
\caption{Hardware and software specifications.}
\label{tab:hardware-spec}
\begin{tabularx}{\columnwidth}{@{}lX@{}}
\toprule
\textbf{Component} & \textbf{Specification} \\
\midrule
GPU & $1 \times$ NVIDIA A800 (80GB HBM2e) \\
VRAM & $80$ GB \\
Host Memory & $128$ GB DDR4 (suggested) \\ 
\midrule
Operating System & Ubuntu 22.04 LTS \\
CUDA Version & 13.0 \\
Deep Learning Framework & PyTorch 2.9.0 \\
Inference Engine & vLLM v0.13.0 \\
\bottomrule
\end{tabularx}
\end{table}

\section{Benchmark Details}
\label{sec:benchmark_details}

Our evaluation encompasses five code generation benchmarks with a total of 836 programming problems. The detailed composition of our test suite is as follows:

\begin{itemize}
    \item \textbf{HumanEval}: Contains 164 programming problems designed to evaluate code generation capabilities on diverse algorithmic tasks.
    
    \item \textbf{MBPP (Mostly Basic Python Problems)}: Comprises 257 problems focusing on basic Python programming skills and common coding patterns.
    
    \item \textbf{LiveCodeBench}: Utilizes the official \texttt{code\_generation\_lite} versions for both v5 and v6 releases:
    \begin{itemize}
        \item LiveCodeBench v5 lite: 167 problems
        \item LiveCodeBench v6 lite: 175 problems
    \end{itemize}
    
    \item \textbf{ICPC}: A subset of OJBench containing 73 competitive programming problems from International Collegiate Programming Contest scenarios.
\end{itemize}

In total, our comprehensive benchmark suite contains $164 + 257 + 167 + 175 + 73 = 836$ programming problems across different difficulty levels and problem domains, providing a thorough evaluation of code generation performance across various coding challenges and paradigms.

\section{Compute Efficiency Analysis}
\label{sec:compute_efficiency_analysis}

\begin{table}[tb!]
\centering
\caption{Compute efficiency comparison on LCB V5 using Qwen3-4B-Instruct-2507. All methods use the same evaluation budget (20 iterations, $k{=}3$, max 2048 tokens). Reported time and FLOPs include full end-to-end cost (forward + backward + search).}
\label{tab:compute_efficiency}
\setlength{\tabcolsep}{3pt}
\renewcommand{\arraystretch}{0.95}
\footnotesize
\begin{tabular}{lccc}
\toprule
\textbf{Method} & \textbf{Acc. (\%)} & \textbf{Avg Time} & \textbf{Est. FLOPs} \\
\midrule
Best-of-N & 43.61 & 36.4 s & 3.02 PF \\
ToT & 43.11 & 96.8 s & 5.94 PF \\
AB-MCTS & 39.52 & 109.7 s & 6.72 PF \\
RethinkMCTS & 49.83 & 101.3 s & 6.08 PF \\
\midrule
TTRL & 48.61 & 84.6 s & 5.96 PF \\
TTT-Discover & 51.38 & 78.9 s & 5.21 PF \\
\midrule
\rowcolor{red!10}
\textbf{PoT (Ours)} & \textbf{57.49} & \textbf{61.7 s} & \textbf{3.86 PF} \\
\bottomrule
\end{tabular}
\end{table}

Table~\ref{tab:compute_efficiency} compares the compute efficiency of representative methods on LCB V5. Despite including the full backward-pass overhead for LoRA updates, PoT achieves the best accuracy (57.49\%) while consuming only 3.86 PFLOPs per problem---less than all search-based and TTT baselines except Best-of-N. This demonstrates that gradient-driven policy evolution effectively prunes the search space, converting backward computation into higher-precision reasoning rather than wasted exploration.

\section{Additional Results on Thinking Models}
\label{sec:thinking_models}

\begin{table}[tb!]
\centering
\small
\caption{Results for Qwen3 Thinking variants on LCB V6 benchmark.}
\label{tab:thinking_lcbv6}
\setlength{\tabcolsep}{4pt}
\begin{tabular}{@{}l|c@{}}
\toprule
Method & LCB V6 \\
\midrule

\multicolumn{2}{@{}l}{\textit{\textbf{Qwen3-4B-Thinking-2507}}} \\
$~~~~$+zero-shot & 55.20 \\
\rowcolor{red!10}
$~~~~$+PoT (Ours) & \textbf{63.70} \\

\midrule

\multicolumn{2}{@{}l}{\textit{\textbf{Qwen3-30B-Thinking-2507}}} \\
$~~~~$+zero-shot & 76.80 \\
\rowcolor{red!10}
$~~~~$+PoT (Ours) & \textbf{84.90} \\

\bottomrule
\end{tabular}
\end{table}

To further examine whether PoT complements models with built-in long-form reasoning, we evaluate it on two dedicated thinking models: Qwen3-4B-Thinking-2507 and Qwen3-30B-Thinking-2507.

As shown in Table~\ref{tab:thinking_lcbv6}, PoT consistently improves both models.
On Qwen3-4B-Thinking-2507, PoT increases LCB V6 accuracy from 55.20 to 63.70 (+8.50 points).
On Qwen3-30B-Thinking-2507, accuracy improves from 76.80 to 84.90 (+8.10 points).

These gains indicate that PoT is complementary to native reasoning capabilities rather than a substitute.
Even when the base model already performs extended multi-step reasoning, test-time policy evolution remains beneficial by refining reasoning trajectories using environment feedback and adapting the policy to the current instance.

\end{document}